\documentclass{article}

\PassOptionsToPackage{numbers, compress}{natbib}

\usepackage[final]{tackling_climate_workshop_style}

\usepackage[utf8]{inputenc} 
\usepackage[T1]{fontenc}    
\usepackage{hyperref}       
\usepackage{url}            
\usepackage{booktabs}       
\usepackage{amsfonts}       
\usepackage{nicefrac}       
\usepackage{microtype}      

\usepackage{multirow}
\usepackage{enumitem}
\usepackage{amsmath}
\usepackage{graphicx}
\usepackage{svg}

\usepackage{algorithm}
\usepackage{algpseudocode}

\usepackage{subcaption}
\usepackage{booktabs}

\usepackage{lscape} 

\usepackage{siunitx} 
\title{FedRAIN-Lite: Federated Reinforcement Algorithms for Improving Idealised \\ Numerical Weather and Climate Models}

\author{%
\parbox{\textwidth}{
\centering
Pritthijit Nath$^{1}$ \quad Sebastian Schemm$^{1}$ \quad Henry Moss$^{1, 2}$ \quad Peter Haynes$^{1}$ \\ \vspace{0.1cm} Emily Shuckburgh$^{3}$ \quad\quad\quad\quad\quad\quad\quad\quad Mark Webb$^{4}$} \vspace{0.2cm} \\ 
$^1$ Department of Applied Mathematics and Theoretical Physics, University of Cambridge\\
$^2$ School of Mathematical Sciences, Lancaster University \\
$^3$ Department of Computer Science and Technology, University of Cambridge\\
$^4$ Met Office Hadley Centre\\
\texttt{\{pn341,ss3299,hm493,phh1,efs20\}@cam.ac.uk}; \texttt{mark.webb@metoffice.gov.uk}
}

\newif\ifshowcontent
\newif\ifnotshowcontent

\showcontenttrue 
\notshowcontentfalse

\begin{document}
\setcitestyle{square}

\maketitle

\begin{abstract}
Sub-grid parameterisations in climate models are traditionally static and tuned offline, limiting adaptability to evolving states. This work introduces \textbf{FedRAIN-Lite}, a federated reinforcement learning (FedRL) framework that mirrors the spatial decomposition used in general circulation models (GCMs) by assigning agents to latitude bands, enabling local parameter learning with periodic global aggregation. Using a hierarchy of simplified energy-balance climate models, from a single-agent baseline (\texttt{ebm-v1}) to multi-agent ensemble (\texttt{ebm-v2}) and GCM-like (\texttt{ebm-v3}) setups, we benchmark three RL algorithms under different FedRL configurations. Results show that Deep Deterministic Policy Gradient (DDPG) consistently outperforms both static and single-agent baselines, with faster convergence and lower area-weighted RMSE in tropical and mid-latitude zones across both \texttt{ebm-v2} and \texttt{ebm-v3} setups. DDPG’s ability to transfer across hyperparameters and low computational cost make it well-suited for geographically adaptive parameter learning. This capability offers a scalable pathway towards high-complexity GCMs and provides a prototype for physically aligned, online-learning climate models that can evolve with a changing climate. Code accessible at %
\ifshowcontent{
\url{https://github.com/p3jitnath/climate-rl-fedrl}.}\fi
\ifnotshowcontent{\url{https://anonymous.4open.science/r/climate-rl-fedRL-B614}.}\fi
\end{abstract}

\section{Introduction}

Climate models are indispensable for understanding the Earth’s many interacting systems, from atmospheric circulation to the hydrological cycle, and play a central role in forecasting weather and projecting future climate impacts. However, their predictive skill is often limited by uncertainties arising from static sub-grid parameterisations of unresolved processes, traditionally tuned offline against observations using expensive, ad-hoc experiments~\cite{hourdin_art_2017, rasp_deep_2018}. This tuning bottleneck often inhibits adaptability to state-dependent variability within the system. Emerging online learning methods, such as Ensemble Kalman Inversion (EnKI)~\cite{iglesias_ensemble_2013}, offer a principled and computationally efficient alternative by casting it as a Bayesian inverse problem, successfully applied to convection schemes in idealised general circulation models (GCMs)~\cite{dunbar_calibration_2021}. Reinforcement learning (RL)~\cite{sutton_reinforcement_1998}, one of the key drivers behind recent advances in large language models (LLMs)~\cite{deepseek-ai_deepseek-r1_2025}, has also shown promise in idealised climate settings, enabling models to iteratively learn parameterisation components by interacting with the climate system itself~\cite{nath_rain_2024}. These recent approaches represent a shift toward adaptive, data-informed parameterisation strategies that can respond to distributional changes over time such as those rising from natural variability or externally forced trends such as global warming.

While Nath et al.~\cite{nath_rain_2024} demonstrated RL’s potential in idealised models, their setups lacked spatial decomposition and treated the system as a whole, limiting scalability and regional adaptivity. In contrast, operational GCMs routinely use spatial decomposition for both physics and computations. Embracing this paradigm, this work introduces a federated reinforcement learning (FedRL) framework~\cite{jin_federated_2022}, which we term \textbf{FedRAIN‑Lite}, where ``federated" refers to the use of multiple agents assigned to distinct latitude bands that learn local policies independently, while synchronising periodically via global aggregation to stabilise training and enable knowledge transfer across bands. This multi‑agent setup not only reduces optimisation complexity within each region but also mirrors real‑world model architectures, enabling both faster convergence and geographically adaptive skill.

As running full-scale GCMs can be computationally expensive and challenging to interpret, we explore a hierarchy of idealised energy balance models (EBMs)~\cite{budyko_effect_1969, sellers_global_1969, north_theory_1975}, where the proposed \textbf{FedRAIN-Lite} framework significantly improves training stability and skill, particularly in tropical and mid-latitude zones, relative to a non-federated approach. Among three RL algorithms tested: Deep Deterministic Policy Gradient (DDPG)~\cite{lillicrap_continuous_2019}, Twin-delayed DDPG (TD3)~\cite{fujimoto_addressing_2018}, and Truncated Quantile Critics (TQC)~\cite{kuznetsov_controlling_2020} (summaries in Appendix~\ref{app:rl-algorithm-summaries}), DDPG emerges as the most {hyperparameter-robust} and computationally efficient candidate, achieving consistent performance gains in both single-agent and federated multi-agent settings. Compared to static baselines and single-agent global RL models, DDPG under federated coordination converges faster and generalises better across latitude bands. 

The key contributions of this work are:
\begin{enumerate}
    \item \textbf{A novel application of FedRL} to climate model parameterisation, using spatial decomposition schemes that mirror the structure of operational GCMs.
    \item \textbf{Systematic benchmarking} of three RL algorithms (DDPG, TD3, TQC) across a hierarchy of idealised EBMs, spanning single-agent and multi-agent configurations with increasing physical complexity.
    \item \textbf{Demonstration of DDPG’s robustness, scalability, and skill}, showing that it consistently achieves strong performance across decompositions and coordination strategies, making it a practical and an efficient baseline for geographically adaptive climate parameter learning.
\end{enumerate}

\section{Methodology}

\enlargethispage{\baselineskip}

\subsection{Background}

In numerical weather and climate models, key unresolved processes such as radiation, convection, and turbulence are represented through parameterisations, which are simplified functional forms with fixed or empirically tuned coefficients. These parameters are typically calibrated offline through expensive trial-and-error simulations or derived from theoretical considerations (often relying on highly idealised assumptions), and thus lack adaptability and can degrade performance under evolving or unseen climate states.

RL offers a compelling alternative by framing parameterisation as a sequential decision process, where an agent learns a control policy that dynamically adjusts parameters based on the evolving model state. Recent work demonstrates RL’s growing impact across science, from fusion plasma stabilisation to environmental management and fluid control~\cite{degrave_magnetic_2022, mole_reinforcement_2025, seo_avoiding_2024, yu_reinforcement_2025, chapman_bridging_2023, cai_reinforcement_2025}. Conventional ML approaches for climate model calibration often lack this feedback-driven adaptability inherent to RL. Moreover, existing RL frameworks typically treat the model as a single unit, neglecting spatial heterogeneity and regional dynamics. This motivates a decentralised approach that better reflects the modular and geographically decomposed design of real-world GCMs.

\subsection{Budyko–Sellers Energy Balance Model}

The Budyko--Sellers EBM~\cite{budyko_effect_1969, sellers_global_1969, north_theory_1975} is a latitudinally resolved idealised climate model that simulates the zonal-mean surface temperature \( T_s(\phi) \) as a function of latitude \( \phi \). It represents the balance between absorbed solar radiation, outgoing longwave radiation, and meridional heat transport (using a downgradient diffusion assumption), governed by the following equation:


\begin{equation}
C(\phi)\frac{\partial T_s}{\partial t} = 
\underbrace{(1 - \alpha(\phi))Q(\phi)}_{\text{absorbed shortwave}} 
- \underbrace{(A + B T_s)}_{\text{longwave cooling}} 
+ \underbrace{\frac{D}{\cos\phi} \frac{\partial}{\partial \phi} \left( \cos\phi \frac{\partial T_s}{\partial \phi} \right)}_{\text{diffusive transport}}
\label{eq:methods-ebm-equation}
\end{equation}


where \( C(\phi) \) is the effective heat capacity, \( \alpha(\phi) \) the surface albedo, \( Q(\phi) \) the insolation, \( A \) and \( B \) the outgoing longwave radiation (OLR) coefficients, and \( D \) the meridional heat transport parameter, typically treated as constants chosen to match observations. Mathematical details on equilibria and instabilities are discussed in Appendix~\ref{app:ebm-equilibria-instabilities}.

The model is discretised into 96 latitude bands, enabling numerical simulation of temperature dynamics across the globe. The OLR parameters \( A \) and \( B \), are the primary targets for optimisation in this work. In the RL setting, these coefficients are treated as learnable policy outputs, with agents optimising them to reduce the temperature error relative to a prescribed climatological target, allowing for spatially adaptive correction policies.

\subsection{climateRL Environments}

For training RL agents, we construct three climateRL environments (schematics in Appendix~\ref{app:climaterl-ebm-schematics}), based on the Budyko--Sellers EBM, increasing in spatial complexity and progressively aligning with real GCM design, as explained below.

{\texttt{ebm-v1}} extends the single-agent RL setup from Nath et al.~\cite{nath_rain_2024}, where a global agent observes the full zonal-mean temperature profile \( T_s(\phi) \) and learns to modulate radiative parameters \( A \) and \( B \) adaptively per latitude to minimise the mean squared error against a target climatology (e.g.. reanalysis). This serves as a single agent centralised baseline without spatial decomposition or regional specialisation.

{\texttt{ebm-v2}} introduces spatial decomposition by assigning latitude bands (grouped into two or six regions) to separate agents. Each agent receives the full temperature profile as input but optimises a region-specific reward. FedRL ensures coordination via periodic global aggregation of local policy networks enabling geographically adaptive learning  while preserving global coherence.

{\texttt{ebm-v3}} mirrors the GCM design by restricting each agent's input to a local temperature slice. This creates a decentralised, partially observed setting closer to real models, where local physics modules operate on region-specific state variables. Like \texttt{ebm-v2}, FedRL is used here with local rewards for global synchronisation.

All climateRL environments are built using climlab~\cite{rose_climlab_2018} and Gymnasium~\cite{towers_gymnasium_2024} , and trained using off-policy algorithms under episodic evaluation. The FedRL setup is implemented via Flower~\cite{beutel_flower_2022}, using synchronous aggregation every \( K \) episodes (e.g., \texttt{fed05}, \texttt{fed10}) to balance trade-offs between local adaptation and global synchronisation.

\section{Results}

\vspace{-0.35cm}
\enlargethispage{1.5\baselineskip}

\begin{figure}[htbp]
    \centering
    \makebox[\textwidth][c]{\includegraphics[height=4cm]{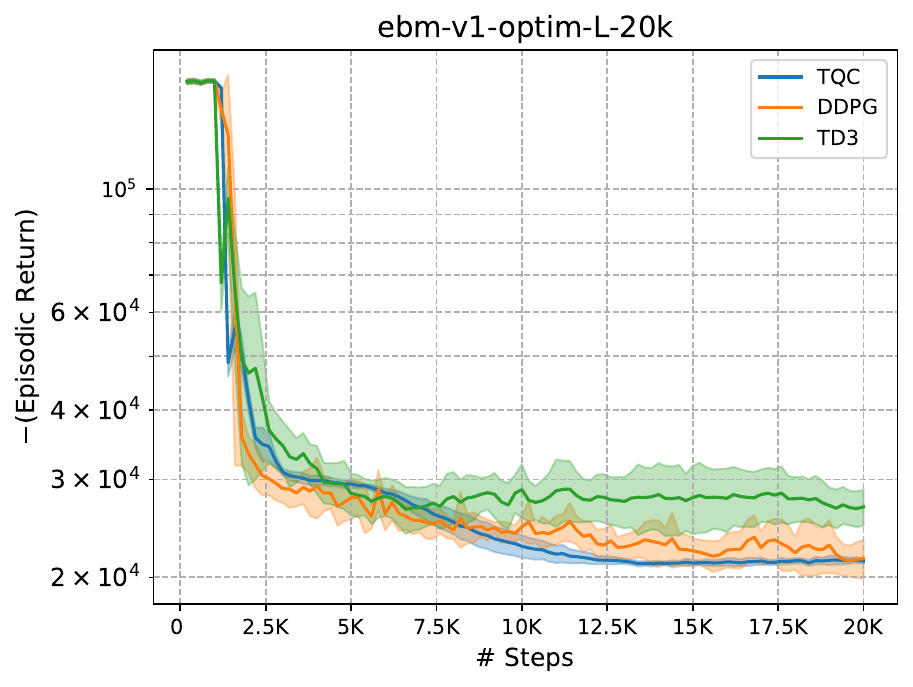}
    \includegraphics[height=4cm]{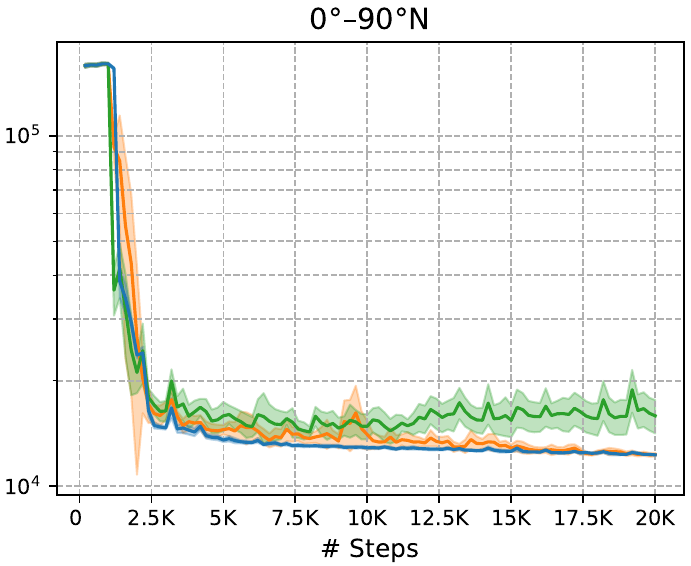}
    \includegraphics[height=4cm]{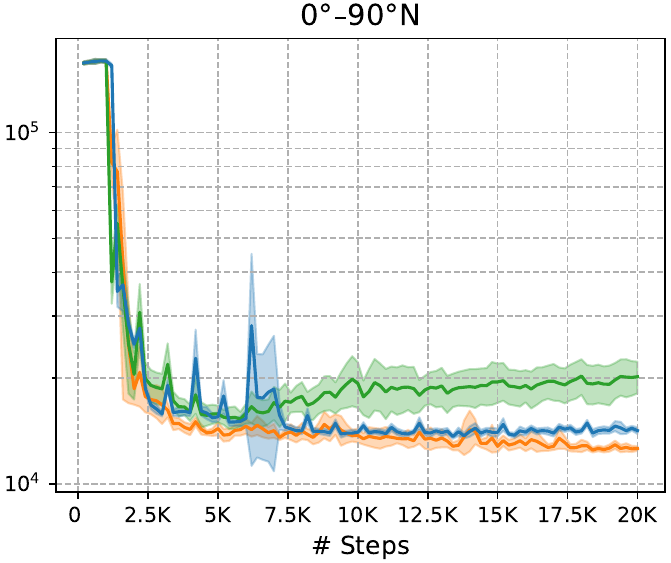}}
    \caption{Episodic return curves (log-scaled) with 95\% spreads over 10 seeds for three RL algorithms: TQC (blue), DDPG (orange), and TD3 (green), across three climateRL environments. Left: \texttt{ebm-v1} (single-agent, global input, global reward and latitude-specific parameters). Middle: \texttt{ebm-v2} (multi-agent FedRL setup with shared global profile input and local rewards). Right: \texttt{ebm-v3} (multi-agent FedRL setup with sliced inputs and local rewards, mirroring GCM-like spatial decomposition).}
    \label{fig:results-ebm-returns}
\end{figure}

\clearpage

\begin{figure}[htbp]
    \centering
    \begin{subfigure}{0.49\textwidth}
        \includegraphics[width=\linewidth]{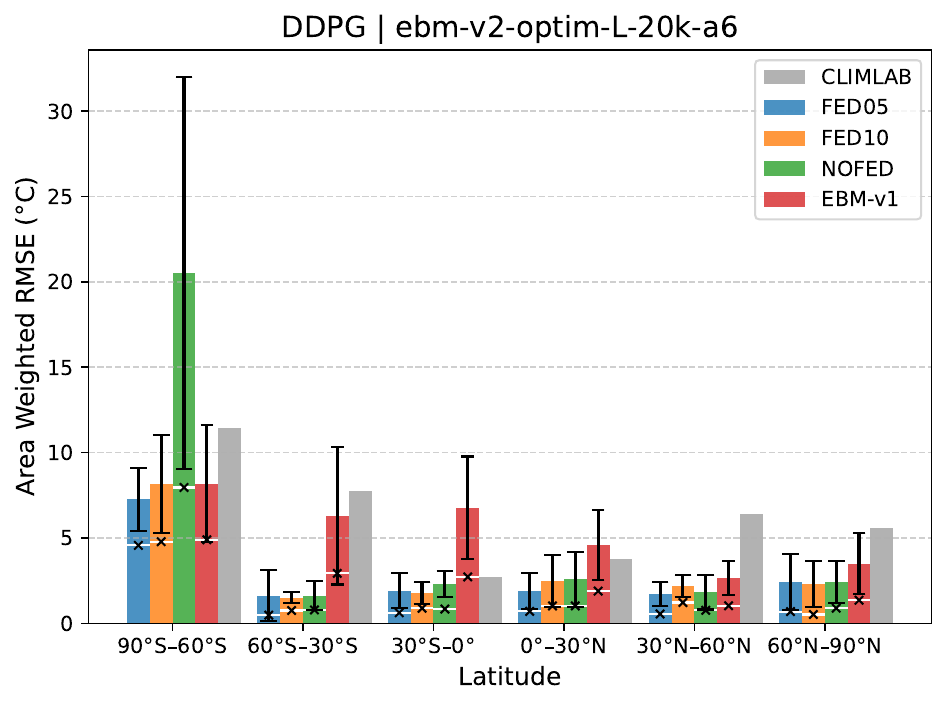}
        \caption{Zonal skill (areaWRMSE) of DDPG in \texttt{ebm-v2}}
        \label{fig:ddpg-zonal-ebm-v3}
    \end{subfigure}
    \hfill
    \begin{subfigure}{0.49\textwidth}
        \includegraphics[width=\linewidth]{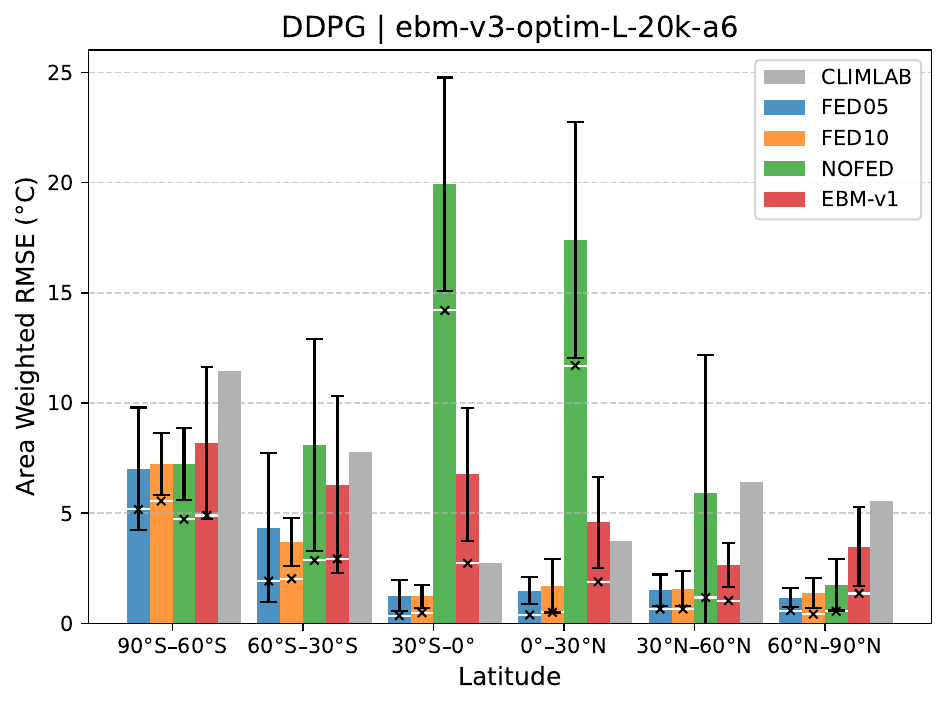}
        \caption{Zonal skill (areaWRMSE) of DDPG in \texttt{ebm-v3}}
        \label{fig:ddpg-zonal-ebm-v2}
    \end{subfigure}
    \caption{Comparison of zonal skill achieved by DDPG under FedRL coordination in \texttt{ebm-v2} and \texttt{ebm-v3}, both using the 6-agent spatial decomposition (\texttt{a6}). Skill is evaluated using areaWRMSE between predicted and reference temperature profiles, averaged with 95\% spreads over 10 seeds. Each subplot reports results for three FedRL schemes: \texttt{fed05}, \texttt{fed10}, \texttt{nofed}, along with single-agent \texttt{ebm-v1} and the static \texttt{climlab} baseline. White horizontal bars with a cross indicate the best-performing seed for each scheme. Both setups adopt the same policy network architecture and hyperparameters as \texttt{ebm-v1}. Detailed skill metrics for all experiments presented in Appendix~\ref{app:ebm-skill-metrics}.\vspace{-0.375cm}}
    \label{fig:results-ebm-ddpg-skill}
\end{figure}

Convergence is significantly faster in the FedRL schemes: \texttt{ebm-v2} and \texttt{ebm-v3} (as can be seen in Figure~\ref{fig:results-ebm-returns}) compared to the single-agent \texttt{ebm-v1} setup. Training curves indicate that most policies stabilise by 2.5k–5k steps in \texttt{ebm-v2/3}, while \texttt{ebm-v1} shows delayed convergence beyond 10k steps. This accelerated training can be attributed to the localised policy learning and reward structures, which reduce the complexity of the optimisation landscape in the decentralised setups. 

In Figure~\ref{fig:results-ebm-ddpg-skill}, across nearly all latitude bands, \texttt{fed05} outperforms both the static baseline and non-federated (\texttt{nofed)} counterparts, achieving significant skill improvements particularly in the tropics (e.g., over 50\% reduction in  area-weighted RMSE (areaWRMSE) in 30°S–0° and 0°–30°N for both \texttt{ebm-v2} and \texttt{ebm-v3}). The gains are more pronounced in \texttt{ebm-v3}, where region-specific inputs likely aid specialisation. In contrast, \texttt{fed10}, while still outperforming \texttt{nofed}, yields higher variance and inconsistent benefits, indicating that frequent aggregation (\texttt{fed05}) is essential for stable coordination. In polar regions, all federated schemes perform comparably to or better than \texttt{ebm-v1}, showcasing that local specialisation assists in challenging loss landscapes with sharp gradients. Even under coarser decomposition (\texttt{a2} in Appendix~\ref{app:ebm-skill-metrics}), DDPG under \texttt{ebm-v2/3} maintains monotonic convergence and achieves low final errors, reinforcing its robustness across spatial setups.

Results here highlight the benefits of regional specialisation via FedRL and confirm DDPG’s hyperparameter robustness to changes in reward structure and input resolution (alongside computational efficiency), reinforcing its suitability for GCM-style architectures. Additional results for TD3 and TQC are provided in Appendix~\ref{app:td3-tqc-additional-results}. While occasionally competitive, both exhibit higher variance and instability, especially under frequent aggregation or in equatorial and polar regions.


\section{Conclusion}

\enlargethispage{\baselineskip}

This work demonstrates that combining RL with federated learning and spatial decomposition offers a scalable and effective strategy for adaptive climate model parameterisation. By aligning with the spatial decomposition used in GCMs, FedRAIN-Lite allows regional agents to learn locally specialised corrections while preserving global coordination. Among the methods evaluated, DDPG consistently achieves stable convergence, low zonal errors, and strong generalisation across both single- and multi-agent setups. By producing region-specific parameter adjustments, the framework also supports interpretability through physical analysis of learnt policies, offering strong potential for future work. Overall, these results position lightweight RL as a practical bridge from idealised EBMs to operational GCMs, paving the way for more responsive, data-driven parameterisations in future climate change assessments.

\ifshowcontent{
}\fi


\begin{ack}
P. Nath was supported by the \href{https://ai4er-cdt.esc.cam.ac.uk/}{UKRI Centre for Doctoral Training in Application of Artificial Intelligence to the study of Environmental Risks} [EP/S022961/1]. Mark Webb was supported by the Met Office Hadley Centre Climate Programme funded by DSIT.
\end{ack}

\bibliographystyle{vancouver}
\bibliography{bibliography_zotero}

\appendix
\renewcommand\thesection{Appendix \Alph{section}}
\renewcommand\thesubsection{\Alph{section}.\arabic{subsection}} 
\renewcommand\thetable{\Alph{section}.\arabic{table}}  
\renewcommand\thefigure{\Alph{section}.\arabic{figure}}  
\setcounter{table}{0}
\setcounter{figure}{0}

\clearpage

\begin{landscape}

\section{Additional Background}
\subsection{RL Algorithm Summaries}
\label{app:rl-algorithm-summaries}

\begin{table}[!h]
\centering
\small
\caption{Four point summaries of the DDPG, TD3 and TQC}
\vspace{0.1cm}
\label{tbl:methods-exp-setup-rl-algos}
\begin{tabular}{lp{12.5cm}}
\toprule
\textbf{Algorithm} & \textbf{Properties} \\ \midrule
Deep Deterministic Policy Gradient (DDPG)~\cite{lillicrap_continuous_2019} & 
1. Off-policy actor-critic algorithm that extends DPG~\cite{silver_deterministic_2014} using deep function approximators and additional stabilisation mechanisms. \newline
2. Introduces experience replay and target networks with soft target updates to decorrelate samples and improve training stability. \newline
3. Actor and critic networks are both updated similar to DPG. \newline
4. Overestimation bias and sensitivity to exploration noise often limit performance unless mitigated by design changes (e.g. TD3). \\ \midrule

Twin Delayed DDPG (TD3)~\cite{fujimoto_addressing_2018} & 
1. Off-policy actor-critic method designed to reduce the overestimation bias observed in DDPG. \newline
2. Uses a double critic architecture where the minimum of two Q-value estimates is used for critic updates. \newline
3. Actor is updated less frequently than the critics, and target networks are softly updated to reduce update variance. \newline
4. Injects temporally correlated Gaussian noise into the target actions to promote exploration in continuous action spaces. \\ \midrule

Truncated Quantile Critics (TQC)~\cite{kuznetsov_controlling_2020} & 
1. Off-policy actor-critic algorithm that builds on SAC~\cite{haarnoja_soft_2018} with a distributional critic using quantile regression. \newline
2. Models the full distribution of returns \( Z(s,a) \) as quantiles \( \{ \tau_i \} \), capturing uncertainty and reducing bias. \newline
3. Discards the top \( k \) quantiles before computing target values to avoid overestimation from outlier returns. \newline
4. Provides a smooth and robust training signal for distributional value estimation by minimising the quantile Huber loss:
\[
\mathcal{L}_\tau = \frac{1}{N} \sum_{i=1}^N \rho_\kappa(\tau_i - y)
\]
where \( \tau_i \) is the predicted \( i \)-th quantile, \( y \) is the target return, \( \rho_\kappa \) denotes the Huber loss with threshold \( \kappa \), and \( N \) is the number of quantile estimates used in training. \\

\bottomrule
\end{tabular}
\end{table}

\end{landscape}

\clearpage

\subsection{EBM Equilibria and Instabilities}
\label{app:ebm-equilibria-instabilities}

\paragraph{Multiple Equilibria and Climate Tipping.}
The Budyko-Sellers EBM can admit multiple steady-state solutions due to the non-linear dependence of albedo on temperature. To understand this, we consider a spatially averaged version of the model, where the diffusion term is neglected (\(D = 0\)) and all quantities are averaged over latitudes. The energy balance equation in Eq.~\ref{eq:methods-ebm-equation} reduces to an ordinary differential equation:
\begin{equation}
C \frac{dT}{dt} = (1 - \alpha(T)) Q - (A + B T)
\end{equation}
At steady state, \( \frac{dT}{dt} = 0 \), we solve:
\begin{equation}
(1 - \alpha(T)) Q = A + B T
\end{equation}
If \( \alpha(T) \) is a smooth or piecewise function with a sharp transition around the freezing point \( T_c \), this equation may admit more than one root. In particular:
\begin{itemize}
    \item A warm stable equilibrium: low albedo (e.g., open ocean), high \( T \)
    \item A cold stable equilibrium: high albedo (e.g., ice-covered), low \( T \)
    \item An intermediate unstable solution separating the two
\end{itemize}
This structure forms an \textit{S-shaped} bifurcation diagram, where gradual changes in solar forcing \( Q \) or feedback strength can lead to sudden jumps between climate states, a phenomenon known as climate tipping. The multiplicity of solutions results from the positive ice-albedo feedback that amplifies temperature perturbations.

\paragraph{Linear Stability of Warm and Cold Equilibria.}
To assess the stability of a given steady-state temperature profile \( T^\star(\phi) \), we linearise the full equation about \( T^\star \). Let \( T(\phi, t) = T^\star(\phi) + \delta T(\phi, t) \), where \( \delta T \) is a small perturbation. Substituting into Eq.~\ref{eq:methods-ebm-equation} and retaining only linear terms gives:
\begin{equation}
C(\phi) \frac{\partial \delta T}{\partial t} = -B \delta T + \frac{D}{\cos \phi} \frac{\partial}{\partial \phi} \left( \cos \phi \frac{\partial \delta T}{\partial \phi} \right)
\end{equation}
or, in transformed coordinates \( x = \sin \phi \):
\begin{equation}
C(x) \frac{\partial \delta T}{\partial t} = -B \delta T + D \frac{d}{dx} \left( (1 - x^2) \frac{d \delta T}{dx} \right)
\end{equation}
This is a linear partial differential equation in \( \delta T(x, t) \), which can be interpreted as an eigenvalue problem:
\begin{equation}
\lambda \delta T = -\frac{B}{C(x)} \delta T + \frac{D}{C(x)} \mathcal{L}[\delta T]
\quad \text{where} \quad
\mathcal{L}[\delta T] := \frac{d}{dx} \left( (1 - x^2) \frac{d \delta T}{dx} \right)
\end{equation}
To solve this, we expand \( \delta T(x, t) \) in terms of the eigenfunctions of the Sturm–Liouville operator \( \mathcal{L} \). These eigenfunctions are the Legendre polynomials \( P_n(x) \), which satisfy:
\begin{equation}
\mathcal{L}[P_n] = -n(n+1) P_n(x)
\end{equation}
Substituting into the eigenvalue equation, we obtain:
\begin{equation}
\lambda_n = -\frac{B}{C} - \frac{D}{C} n(n+1)
\label{eq:app-methods-rlenvs-ebm-linearised-stability}
\end{equation}
where \( n = 0, 1, 2, \ldots \) and we assume \( C(x) = C \) is constant for simplicity. \\

Each eigenvalue \( \lambda_n \) is strictly negative since both terms are negative, implying that all modes decay exponentially with time. The higher the value of \( n \), the faster the decay, corresponding to the damping of fine-scale spatial features. Hence, the equilibrium is linearly stable. If \( C(x) \) varies with latitude, this eigenvalue spectrum will be modified accordingly, but the sign of the real part remains governed by the relative magnitudes of \( B \), \( D \), and the spatial variation in \( C(x) \). 

\clearpage

The spectrum of eigenvalues \( \{ \lambda_n \} \) provides insight into the stability of the different equilibria:

\begin{itemize}
    \item \textbf{Warm Equilibrium.} At the warm stable branch, the surface temperature \( \overline{T_s}(x) \) remains well above the ice threshold \( T_c \) over most latitudes, resulting in a uniformly low albedo \( \alpha(\phi) \approx \alpha_{\text{water}} \). The radiative damping coefficient \( B \) and diffusivity \( D \) dominate the dynamics, ensuring that all eigenvalues \( \lambda_n \) remain strictly negative. This guarantees that all perturbation modes decay exponentially, and the equilibrium is linearly stable.

    \item \textbf{Cold Equilibrium.} In the cold branch, nearly the entire domain is ice-covered, and \( \alpha(\phi) \approx \alpha_{\text{ice}} \) is large. The outgoing radiation \( A + B \overline{T_s} \) is lower due to lower temperatures, but the structure of the eigenvalue spectrum remains similar. Although the temperature sensitivity of albedo becomes small (flat high-albedo state), the overall damping remains strong, and the eigenvalues \( \lambda_n \) are still negative. Thus, this branch is also linearly stable.
\end{itemize}

\paragraph{Instability in the Intermediate Branch.}
Although the linearised eigenvalue spectrum for constant-coefficient EBM in Eq.~\ref{eq:app-methods-rlenvs-ebm-linearised-stability} yields strictly negative eigenvalues, this result assumes that the albedo \( \alpha(x) \) is independent of temperature. In the intermediate equilibrium, however, the steady-state temperature \( \overline{T}(x) \) lies near the ice–water transition threshold \( T_c \), and hence small perturbations in temperature cause large changes in albedo. 

To account for temperature-dependent albedo in the linear stability analysis, we modify the perturbed form of the EBM:
\begin{equation}
C(x) \frac{\partial \delta T}{\partial t}
= -B \delta T + D \frac{d}{dx} \left( (1 - x^2) \frac{d \delta T}{dx} \right)
- Q(x) \cdot \frac{d\alpha}{dT} \cdot \delta T 
\end{equation}
Here, the albedo perturbation introduces an additional source term through the chain rule:
\begin{equation}
\delta \alpha(x) = \frac{d\alpha}{dT} \cdot \delta T(x)
\end{equation}

Substituting into the linearised energy balance yields a modified restoring term:
\begin{equation}
-B \delta T(x) - Q(x) \cdot \frac{d\alpha}{dT} \cdot \delta T(x)
= - \left( B + Q(x) \cdot \frac{d\alpha}{dT} \right) \delta T(x)
\end{equation}
This implies an \textbf{effective damping} coefficient:
\begin{equation}
B_{\text{eff}}(x) = B + Q(x) \cdot \frac{d\alpha}{dT}
\end{equation}

As a result, the eigenvalue equation becomes:
\begin{equation}
\lambda \delta T = -\frac{B_{\text{eff}}(x)}{C(x)} \delta T
+ \frac{D}{C(x)} \cdot \frac{d}{dx} \left( (1 - x^2) \frac{d \delta T}{dx} \right)
\end{equation}

When \( \frac{d\alpha}{dT} \) is large and negative, typical near the albedo transition temperature \( T_c \), the effective damping term \( B_{\text{eff}}(x) \) can become negative in parts of the domain. If this occurs over a sufficiently wide region in \( x \in [-1, 1] \), the dominant eigenvalue \( \lambda_0 \) may cross zero and become positive, indicating the onset of linear instability. 

This mechanism explains how the intermediate equilibrium, where the climate state is sensitive to ice–albedo feedback near the freezing threshold, is destabilised. In contrast, both the warm and cold equilibria satisfy \( \frac{d\alpha}{dT} \approx 0 \), leading to \( B_{\text{eff}}(x) \approx B > 0 \) and a strictly negative eigenvalue spectrum, ensuring stability.

\clearpage

\subsection{climateRL EBM Schematics}
\label{app:climaterl-ebm-schematics}

\begin{figure}[!h]
    \centering
    \begin{subfigure}{\textwidth}
        \centering
        \includegraphics[width=0.6\linewidth]{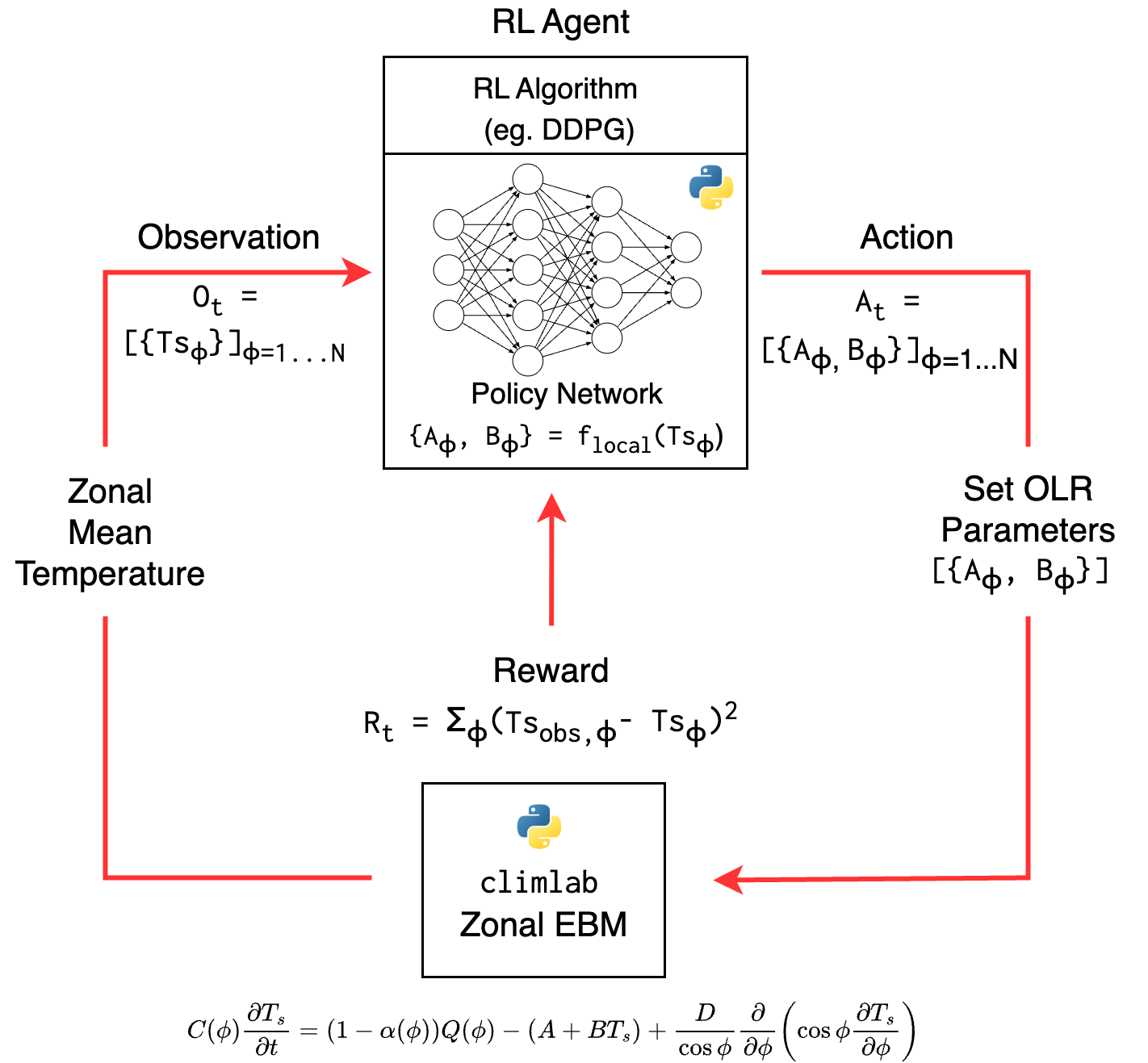}
        \caption{\texttt{ebm-v1} single-agent setup. The global agent observes the full zonal-mean temperature profile and outputs latitude-dependent radiative parameters \( \{A_\phi, B_\phi\} \). Loss from observations are computed over all 96 latitudes.}
        \label{fig:app-ebm-v1-dataflow}
    \end{subfigure}
    \hfill
    \vspace{0.15cm}
    \begin{subfigure}{\textwidth}
        \includegraphics[width=\linewidth]{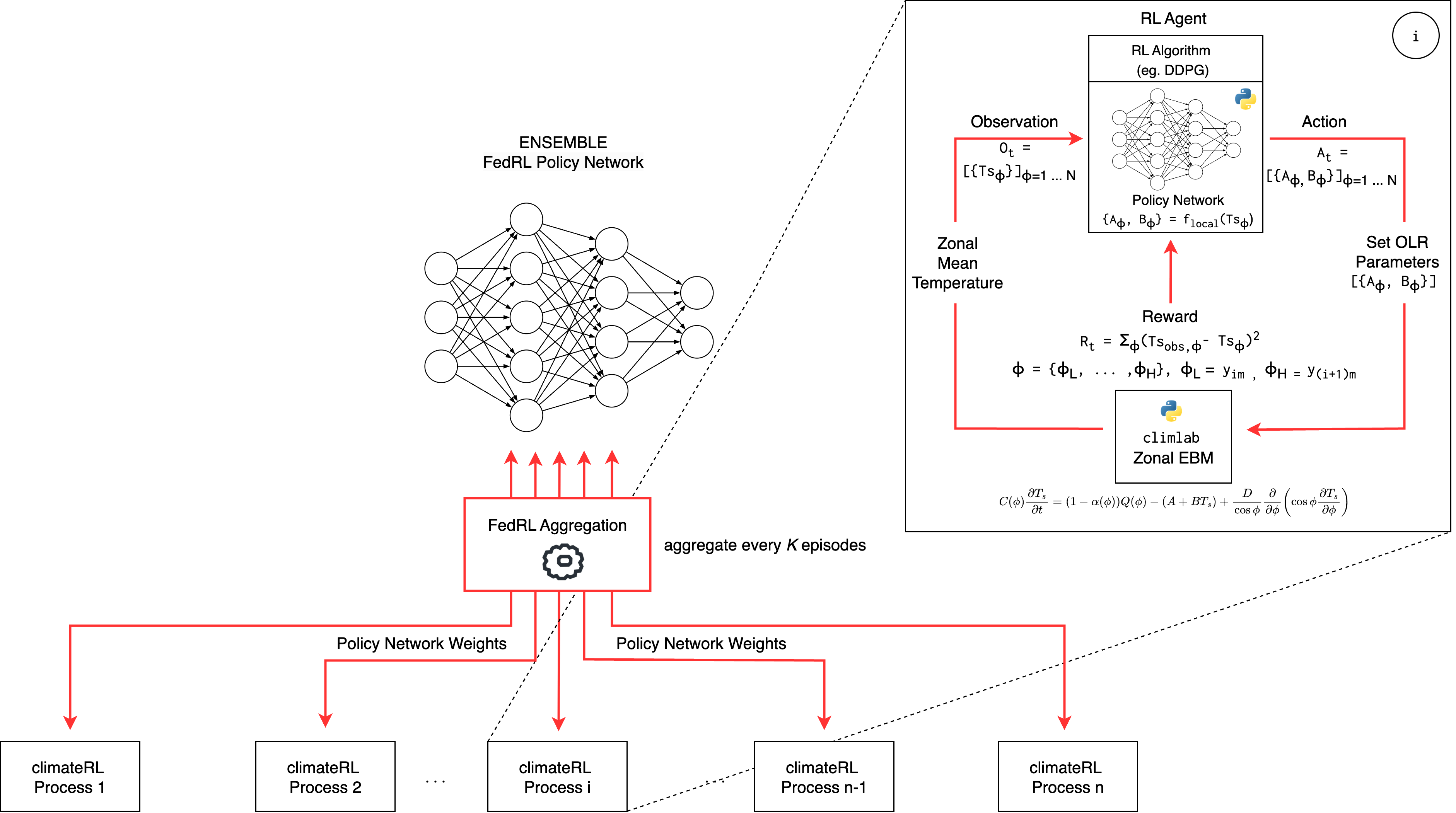}
        \caption{\texttt{ebm-v2} multi-agent ensemble with FedRL
        agents operate on latitude groups with local rewards while receiving the global profile as input. Periodic aggregation every \(K\) episodes synchronises policy weights across \(n\) agents.}
        \label{fig:app-ebm-v2-dataflow}
    \end{subfigure}
    \caption{Schematics for climateRL EBM environments}
    \label{fig:ebm-v123-dataflow}
\end{figure}

\begin{landscape}
\begin{figure}[p]
    \centering
    \ContinuedFloat
    \includegraphics[width=0.9\paperwidth]{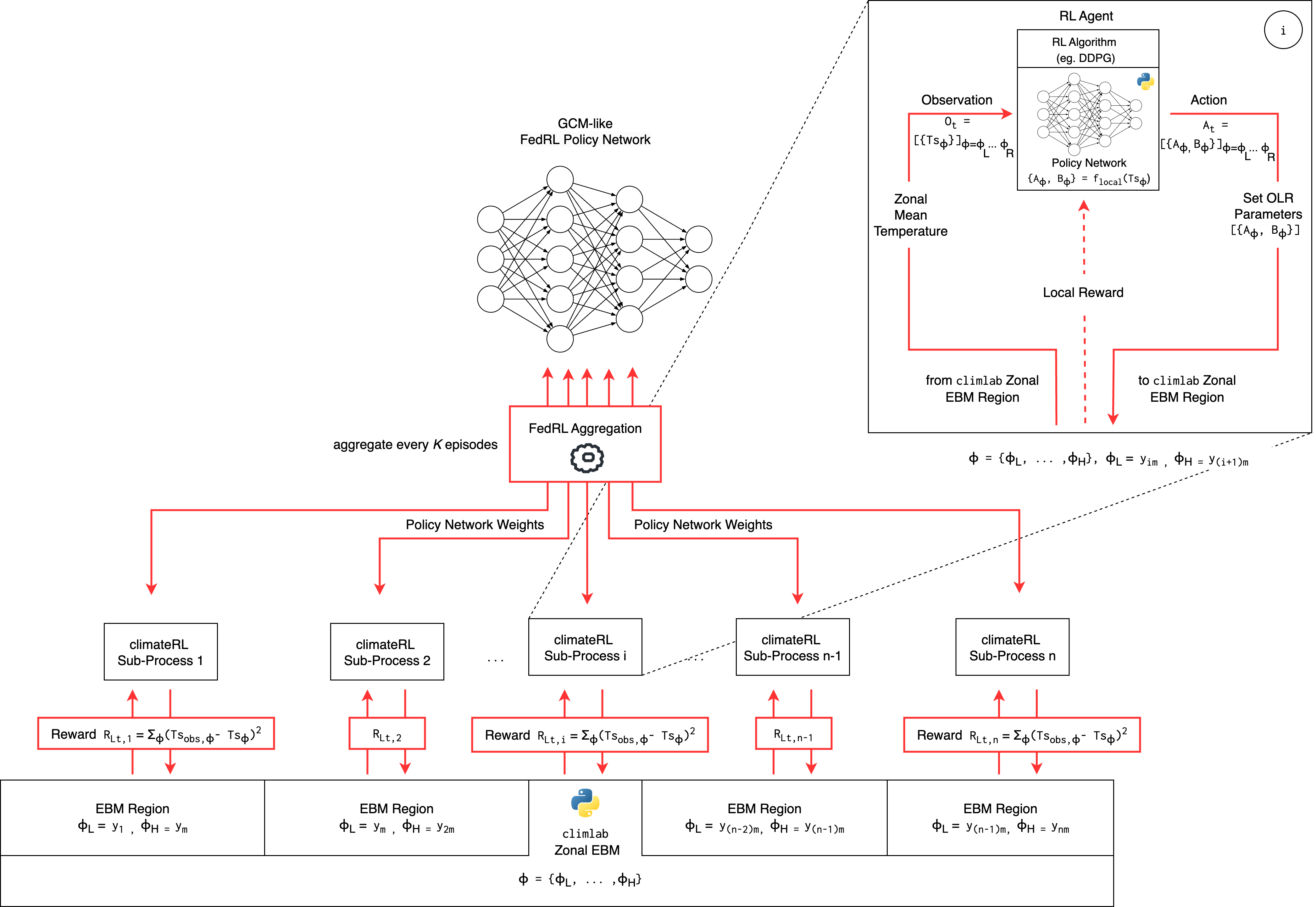}
    \caption{Schematics for GCM-like \texttt{ebm-v3}. FedRL agents observe local temperature slices only (unlike \texttt{ebm-v2}) and optimise local rewards. Local rewards are computed over regions for each agent. Policy weights are aggregated every \(K\) episodes via Flower for FedRL and also contribute towards a global non-local policy.}
    \label{fig:app-ebm-v123-dataflow}
\end{figure}
\end{landscape}

\subsection{RL Algorithm Hyperparameters}
\label{app:rl-algorithm-hyperparameters}

\begin{table}[!h]
\centering
\small
\ttfamily
\caption{Tabular representation of different RL hyperparameters}
\vspace{0.25em}
\label{tbl:tune_parameters}
\begin{tabular}{lp{10cm}l}
\toprule
\textbf{\textrm{Algorithm}} & \textbf{\textrm{Parameter Names}} & \textbf{\textrm{Count}}\\
\midrule
\textrm{DDPG} & learning\_rate, tau, batch\_size, exploration\_noise, policy\_frequency, noise\_clip, actor\_critic\_layer\_size & 7 \\ \midrule
\textrm{TD3} & learning\_rate, tau, batch\_size, policy\_noise, exploration\_noise, policy\_frequency, noise\_clip, actor\_critic\_layer\_size & 8 \\ \midrule
\textrm{TQC} & tau, batch\_size, n\_quantiles, n\_critics, actor\_adam\_lr, critic\_adam\_lr, alpha\_adam\_lr, policy\_frequency, target\_network\_frequency, actor\_critic\_layer\_size & 10 \\ 
\bottomrule
\end{tabular}
\end{table}

\subsection{Experimental Outline}

\begin{figure}[htbp]
    \centering
    \includegraphics[width=0.8\textwidth]{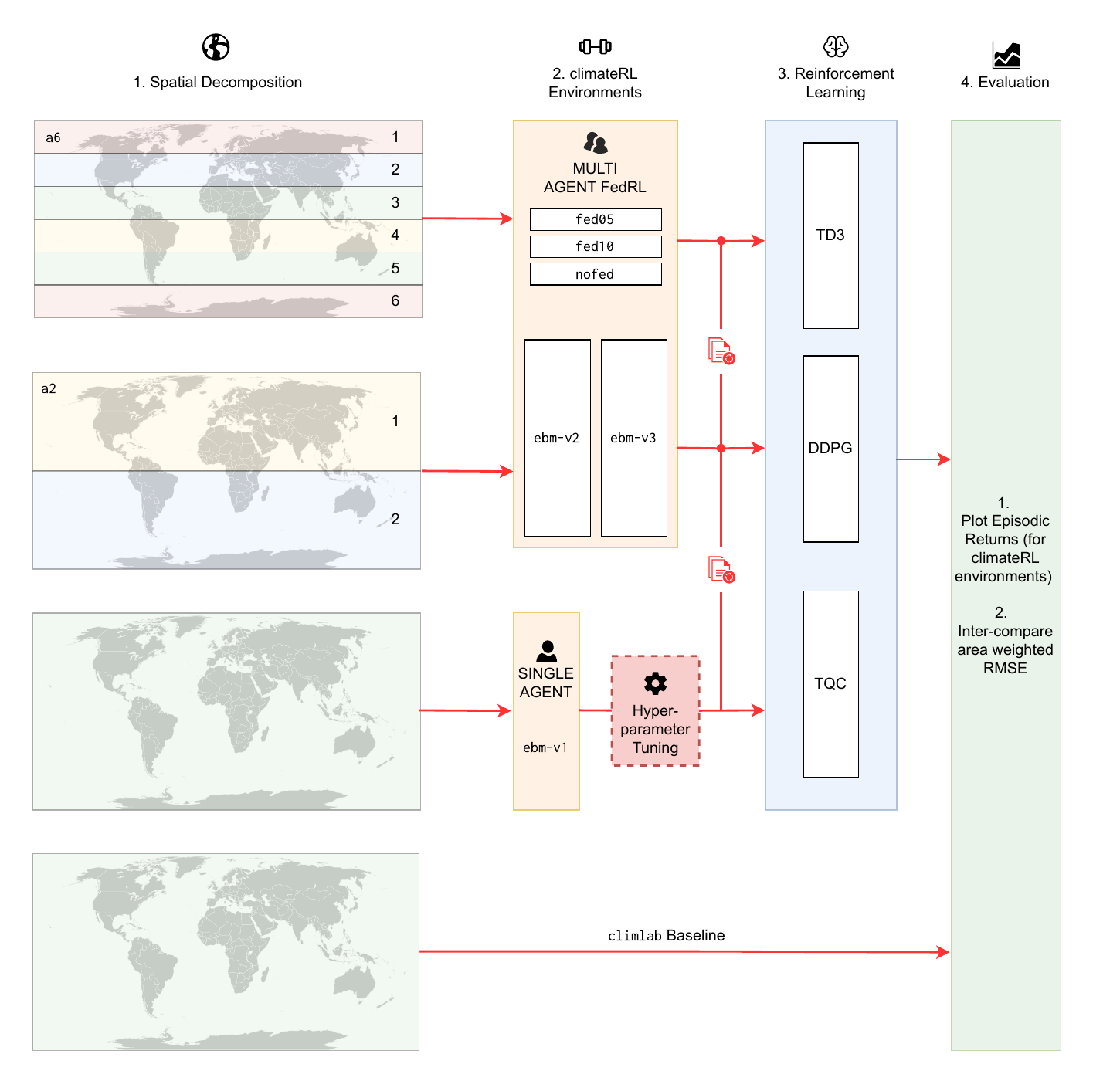}
    \caption{Pipeline for the \texttt{ebm-v1/2/3} experiments. The process begins with configuring the Budyko–Sellers EBM in either single-agent (\texttt{ebm-v1}) or spatially decomposed multi-agent forms (\texttt{ebm-v2}, \texttt{ebm-v3}) using two (\texttt{a2}) or six (\texttt{a6}) regions. Agents are trained with one of three RL algorithms (DDPG, TD3, TQC) under coordination schemes \texttt{fed05}, \texttt{fed10}, or \texttt{nofed}. In multi-agent settings, policies are periodically aggregated via FedRL every \(K\) episodes. Hyperparameters tuned for \texttt{ebm-v1} are transferred over to \texttt{ebm-v2/v3}. Finally trained models are assessed on their training curves and benchmarked against a static \texttt{climlab} baseline, using a skill measure such as areaWRMSE across 30° latitude groups.}
    \label{fig:app-ebm-experiment-flow}
\end{figure}

\clearpage

\subsection{EBM State Evolution}

\begin{figure}[!ht]
    \centering
    \includegraphics[width=\textwidth]{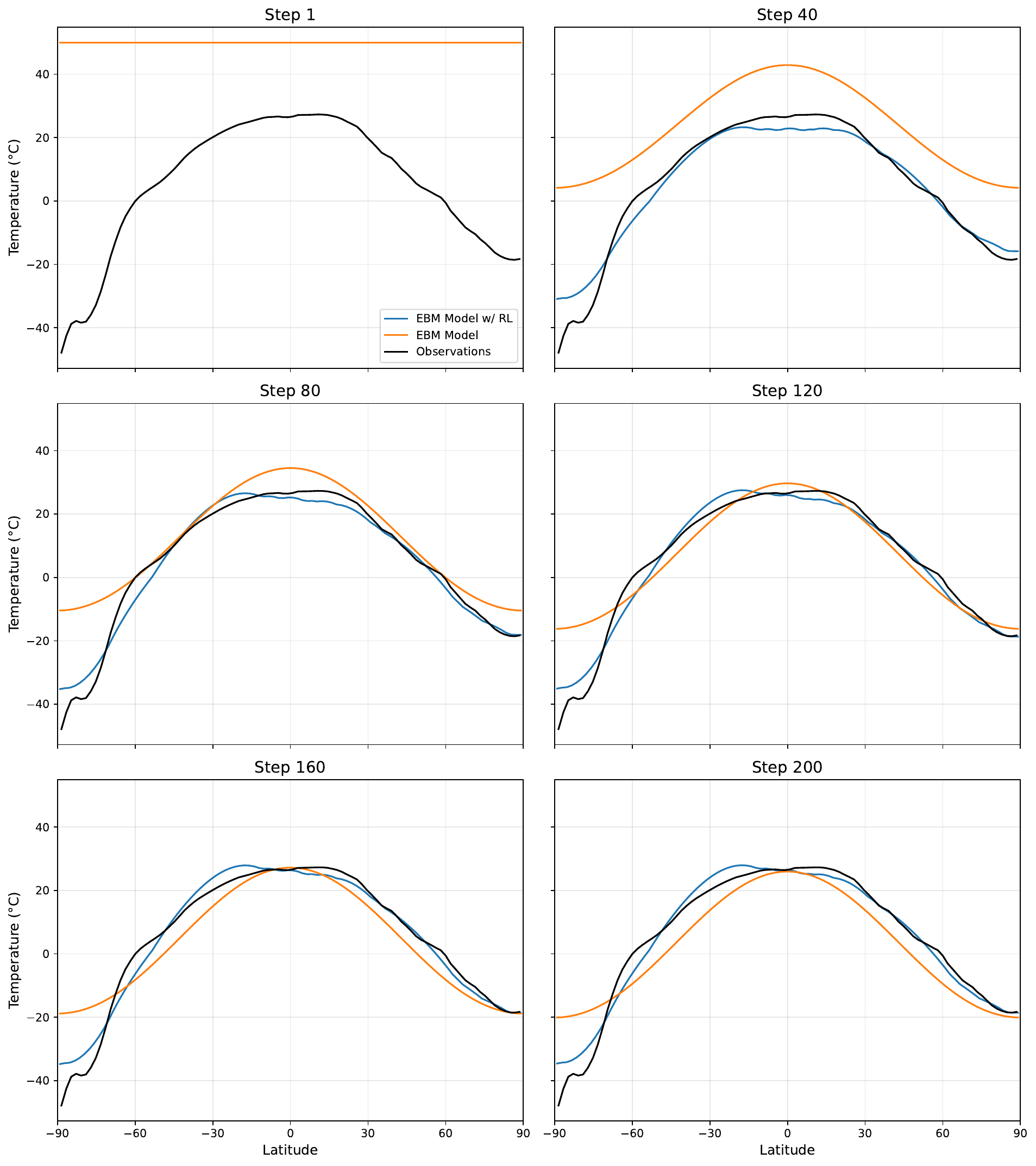}
    \caption{
        Evolution of the zonal-mean surface temperature in the \texttt{ebm-v1} climateRL environment over 200 integration steps. Each panel shows the latitudinal temperature profile at a selected timestep (\(t = 1, 40, 80, 120, 160, 200\)), comparing the DDPG-assisted EBM (blue) with the standard \texttt{climlab} EBM (orange) and reanalysis observations (black). The RL agent dynamically adjusts the OLR parameters \(A\) and \(B\) per latitude, improving temperature representation while maintaining physical consistency.
    }
    \label{fig:ebm_state_evolution}
\end{figure}

\clearpage

\begin{landscape}

\section{Additional Results}
\subsection{Skill Metrics}
\label{app:ebm-skill-metrics}

\enlargethispage{4\baselineskip}

\begin{table}[!h]
\centering
\small
\caption{Zonal‑band errors for \texttt{ebm-v2-optim-L-20k-a2}. Each subtable reports mean ± std and relative gain \% versus \texttt{ebm-v1} for three regimes \texttt{fed05}, \texttt{fed10} and \texttt{nofed}, along with a comparison against the static baseline \texttt{climlab}}
\setlength{\tabcolsep}{4pt}
\captionsetup[sub]{font=small}
\begin{minipage}[t]{\textwidth}
\centering
\subcaption{DDPG}
\begin{tabular}{l c cc cc cc c}
\toprule
 & \multirow{2}{*}{\texttt{climlab}} & \multicolumn{2}{c}{\texttt{fed05}} & \multicolumn{2}{c}{\texttt{fed10}} & \multicolumn{2}{c}{\texttt{nofed}} & \multirow{2}{*}{\texttt{ebm-v1}} \\
 &  & Mean ± Std & Gain \% & Mean ± Std & Gain \% & Mean ± Std & Gain \% &  \\
\midrule
90°S–60°S & 11.453 & 5.11 ± 0.533 & 37.550 & 6.14 ± 0.964 & 25.040 & 16.17 ± 14.718 & -97.490 & 8.19 ± 3.433 \\
60°S–30°S & 7.768  & 3.08 ± 1.155 & 51.030 & 3.39 ± 0.775 & 46.080 & 10.30 ± 10.647 & -63.620 & 6.30 ± 4.019 \\
30°S–0°   & 2.730  & 3.46 ± 1.984 & 48.890 & 3.92 ± 1.642 & 42.080 & 12.24 ± 13.889 & -81.120 & 6.76 ± 3.013 \\
0°–30°N   & 3.746  & 2.80 ± 2.384 & 39.050 & 1.89 ± 1.325 & 58.760 & 2.23 ± 1.246  & 51.460  & 4.59 ± 2.068 \\
30°N–60°N & 6.398  & 2.35 ± 0.866 & 11.350 & 2.26 ± 1.307 & 14.730 & 2.38 ± 1.253  & 10.360  & 2.65 ± 0.997 \\
60°N–90°N & 5.566  & 1.60 ± 0.682 & 54.090 & 1.95 ± 0.995 & 44.060 & 2.49 ± 0.758  & 28.400  & 3.48 ± 1.790 \\
\bottomrule
\end{tabular}
\vspace{0.1cm}
\end{minipage}\hfill
\begin{minipage}[t]{\textwidth}
\centering
\subcaption{TD3}
\begin{tabular}{l c cc cc cc c}
\toprule
 & \multirow{2}{*}{\texttt{climlab}} & \multicolumn{2}{c}{\texttt{fed05}} & \multicolumn{2}{c}{\texttt{fed10}} & \multicolumn{2}{c}{\texttt{nofed}} & \multirow{2}{*}{\texttt{ebm-v1}} \\
 &  & Mean ± Std & Gain \% & Mean ± Std & Gain \% & Mean ± Std & Gain \% &  \\
\midrule
90°S–60°S & 11.453 & 8.00 ± 1.380 & 17.850  & 6.87 ± 0.825 & 29.520  & 7.72 ± 1.371 & 20.750  & 9.74 ± 3.330 \\
60°S–30°S & 7.768  & 7.32 ± 2.327 & -2.870  & 5.51 ± 2.337 & 22.530  & 4.98 ± 1.696 & 29.980  & 7.12 ± 3.800 \\
30°S–0°   & 2.730  & 4.47 ± 1.889 & -17.750 & 4.77 ± 2.313 & -25.480 & 3.49 ± 1.969 & 8.210   & 3.80 ± 1.713 \\
0°–30°N   & 3.746  & 4.03 ± 2.409 & -42.080 & 5.67 ± 3.152 & -100.100& 4.06 ± 2.277 & -43.360 & 2.83 ± 1.247 \\
30°N–60°N & 6.398  & 4.30 ± 2.556 & 14.860  & 4.79 ± 3.138 & 5.170   & 4.39 ± 1.941 & 13.060  & 5.06 ± 2.602 \\
60°N–90°N & 5.566  & 3.54 ± 1.867 & 34.680  & 3.50 ± 1.518 & 35.360  & 5.63 ± 1.948 & -4.030  & 5.42 ± 2.721 \\
\bottomrule
\end{tabular}
\vspace{0.1cm}
\end{minipage}\hfill
\begin{minipage}[t]{\textwidth}
\centering
\subcaption{TQC}
\begin{tabular}{l c cc cc cc c}
\toprule
 & \multirow{2}{*}{\texttt{climlab}} & \multicolumn{2}{c}{\texttt{fed05}} & \multicolumn{2}{c}{\texttt{\texttt{fed10}}} & \multicolumn{2}{c}{\texttt{nofed}} & \multirow{2}{*}{\texttt{ebm-v1}} \\
 &  & Mean ± Std & Gain \% & Mean ± Std & Gain \% & Mean ± Std & Gain \% &  \\
\midrule
90°S–60°S & 11.453 & 8.28 ± 0.896 & 8.550  & 8.13 ± 0.765 & 10.300 & 8.85 ± 1.124 & 2.320  & 9.06 ± 1.549 \\
60°S–30°S & 7.768  & 7.46 ± 1.618 & 6.440  & 7.08 ± 1.382 & 11.120 & 8.24 ± 1.716 & -3.390 & 7.97 ± 1.459 \\
30°S–0°   & 2.730  & 2.83 ± 1.195 & -26.510& 2.34 ± 1.053 & -4.390 & 3.32 ± 1.269 & -48.170& 2.24 ± 0.868 \\
0°–30°N   & 3.746  & 2.30 ± 0.511 & -25.400& 2.19 ± 0.604 & -19.480& 2.49 ± 1.074 & -35.470& 1.84 ± 0.561 \\
30°N–60°N & 6.398  & 0.93 ± 0.222 & 61.460 & 0.92 ± 0.149 & 61.860 & 1.23 ± 0.423 & 49.160 & 2.42 ± 0.706 \\
60°N–90°N & 5.566  & 1.34 ± 0.219 & 42.680 & 1.33 ± 0.352 & 43.140 & 1.44 ± 0.297 & 38.240 & 2.33 ± 0.767 \\
\bottomrule
\end{tabular}
\end{minipage}
\end{table}

\clearpage

\begin{table}[!h]
\centering
\small
\caption{Zonal‑band errors for \texttt{ebm-v2-optim-L-20k-a6}. Each subtable reports mean ± std and relative gain \% versus \texttt{ebm-v1} for three regimes \texttt{fed05}, \texttt{fed10} and \texttt{nofed}, along with a comparison against the static baseline \texttt{climlab}}
\setlength{\tabcolsep}{4pt}
\captionsetup[sub]{font=small}
\begin{minipage}[t]{\textwidth}
\centering
\subcaption{DDPG}
\begin{tabular}{l c cc cc cc c}
\toprule
 & \multirow{2}{*}{\texttt{climlab}} & \multicolumn{2}{c}{\texttt{fed05}} & \multicolumn{2}{c}{\texttt{fed10}} & \multicolumn{2}{c}{\texttt{nofed}} & \multirow{2}{*}{\texttt{ebm-v1}} \\
 &  & Mean ± Std & Gain \% & Mean ± Std & Gain \% & Mean ± Std & Gain \% &  \\
\midrule
90°S–60°S & 11.453 & 7.26 ± 1.845 & 11.340 & 8.17 ± 2.886 & 0.260  & 20.52 ± 11.476 & -150.600 & 8.19 ± 3.433 \\
60°S–30°S & 7.768  & 1.63 ± 1.478 & 74.100 & 1.51 ± 0.324 & 76.020 & 1.62 ± 0.843  & 74.240  & 6.30 ± 4.019 \\
30°S–0°   & 2.730  & 1.92 ± 1.029 & 71.600 & 1.79 ± 0.624 & 73.580 & 2.31 ± 0.776  & 65.870  & 6.76 ± 3.013 \\
0°–30°N   & 3.746  & 1.89 ± 1.032 & 58.750 & 2.49 ± 1.509 & 45.850 & 2.59 ± 1.607  & 43.640  & 4.59 ± 2.068 \\
30°N–60°N & 6.398  & 1.71 ± 0.697 & 35.620 & 2.19 ± 0.643 & 17.530 & 1.81 ± 0.998  & 31.730  & 2.65 ± 0.997 \\
60°N–90°N & 5.566  & 2.43 ± 1.643 & 30.190 & 2.30 ± 1.338 & 34.050 & 2.42 ± 1.224  & 30.670  & 3.48 ± 1.790 \\
\bottomrule
\end{tabular}
\vspace{0.1cm}
\end{minipage}\hfill
\begin{minipage}[t]{\textwidth}
\centering
\subcaption{TD3}
\begin{tabular}{l c cc cc cc c}
\toprule
 & \multirow{2}{*}{\texttt{climlab}} & \multicolumn{2}{c}{\texttt{fed05}} & \multicolumn{2}{c}{\texttt{fed10}} & \multicolumn{2}{c}{\texttt{nofed}} & \multirow{2}{*}{\texttt{ebm-v1}} \\
 &  & Mean ± Std & Gain \% & Mean ± Std & Gain \% & Mean ± Std & Gain \% &  \\
\midrule
90°S–60°S & 11.453 & 7.01 ± 2.553 & 28.000  & 6.04 ± 0.933 & 38.030  & 15.90 ± 9.258 & -63.150  & 9.74 ± 3.330 \\
60°S–30°S & 7.768  & 3.16 ± 1.133 & 55.560  & 3.52 ± 1.276 & 50.510  & 3.52 ± 1.627 & 50.480   & 7.12 ± 3.800 \\
30°S–0°   & 2.730  & 7.68 ± 2.389 & -102.170& 8.16 ± 1.813 & -114.730& 6.10 ± 2.088 & -60.670  & 3.80 ± 1.713 \\
0°–30°N   & 3.746  & 7.27 ± 2.023 & -156.640& 7.63 ± 1.859 & -169.470& 6.05 ± 2.532 & -113.500 & 2.83 ± 1.247 \\
30°N–60°N & 6.398  & 3.92 ± 1.599 & 22.380  & 3.80 ± 0.851 & 24.770  & 3.47 ± 1.360 & 31.460   & 5.06 ± 2.602 \\
60°N–90°N & 5.566  & 2.97 ± 1.614 & 45.090  & 2.48 ± 1.563 & 54.120  & 5.55 ± 5.617 & -2.380   & 5.42 ± 2.721 \\
\bottomrule
\end{tabular}
\vspace{0.1cm}
\end{minipage}\hfill
\begin{minipage}[t]{\textwidth}
\centering
\subcaption{TQC}
\begin{tabular}{l c cc cc cc c}
\toprule
 & \multirow{2}{*}{\texttt{climlab}} & \multicolumn{2}{c}{\texttt{fed05}} & \multicolumn{2}{c}{\texttt{\texttt{fed10}}} & \multicolumn{2}{c}{\texttt{nofed}} & \multirow{2}{*}{\texttt{ebm-v1}} \\
 &  & Mean ± Std & Gain \% & Mean ± Std & Gain \% & Mean ± Std & Gain \% &  \\
\midrule
90°S–60°S & 11.453 & 34.96 ± 7.034 & -285.900 & 28.35 ± 7.956 & -212.890 & 33.90 ± 7.319 & -274.160 & 9.06 ± 1.549 \\
60°S–30°S & 7.768  & 1.68 ± 1.087  & 78.890  & 1.69 ± 1.215  & 78.780  & 1.28 ± 0.393  & 83.920  & 7.97 ± 1.459 \\
30°S–0°   & 2.730  & 1.97 ± 1.814  & 11.850  & 2.25 ± 1.590  & -0.480  & 0.80 ± 0.213  & 64.350  & 2.24 ± 0.868 \\
0°–30°N   & 3.746  & 1.21 ± 0.670  & 34.310  & 1.77 ± 1.363  & 3.430   & 0.75 ± 0.190  & 59.300  & 1.84 ± 0.561 \\
30°N–60°N & 6.398  & 1.97 ± 1.492  & 18.750  & 1.73 ± 0.551  & 28.600  & 1.17 ± 0.330  & 51.920  & 2.42 ± 0.706 \\
60°N–90°N & 5.566  & 30.70 ± 8.534 & -1215.560& 32.88 ± 9.606 & -1308.920& 43.12 ± 14.594& -1747.640& 2.33 ± 0.767 \\
\bottomrule
\end{tabular}
\end{minipage}
\end{table}

\begin{table}[!h]
\centering
\small
\caption{Zonal‑band errors for \texttt{ebm-v3-optim-L-20k-a2}. Each subtable reports mean ± std and relative gain \% versus \texttt{ebm-v1} for three regimes \texttt{fed05}, \texttt{fed10} and \texttt{nofed}, along with a comparison against the static baseline \texttt{climlab}}
\setlength{\tabcolsep}{4pt}
\captionsetup[sub]{font=small}
\begin{minipage}[t]{\textwidth}
\centering
\subcaption{DDPG}
\begin{tabular}{l c cc cc cc c}
\toprule
 & \multirow{2}{*}{\texttt{climlab}} & \multicolumn{2}{c}{\texttt{fed05}} & \multicolumn{2}{c}{\texttt{fed10}} & \multicolumn{2}{c}{\texttt{nofed}} & \multirow{2}{*}{\texttt{ebm-v1}} \\
 &  & Mean ± Std & Gain \% & Mean ± Std & Gain \% & Mean ± Std & Gain \% &  \\
\midrule
90°S–60°S & 11.453 & 6.69 ± 1.766 & 18.240 & 7.52 ± 2.718 & 8.180  & 7.76 ± 2.193 & 5.200  & 8.19 ± 3.433 \\
60°S–30°S & 7.768  & 3.30 ± 1.168 & 47.540 & 4.20 ± 1.272 & 33.290 & 3.98 ± 2.135 & 36.810 & 6.30 ± 4.019 \\
30°S–0°   & 2.730  & 4.84 ± 2.151 & 28.340 & 3.77 ± 2.190 & 44.220 & 3.26 ± 1.873 & 51.720 & 6.76 ± 3.013 \\
0°–30°N   & 3.746  & 2.42 ± 1.786 & 47.180 & 2.96 ± 2.276 & 35.560 & 2.49 ± 1.461 & 45.840 & 4.59 ± 2.068 \\
30°N–60°N & 6.398  & 2.00 ± 0.885 & 24.650 & 2.73 ± 1.056 & -2.980 & 2.67 ± 1.400 & -0.700 & 2.65 ± 0.997 \\
60°N–90°N & 5.566  & 1.96 ± 0.681 & 43.670 & 1.69 ± 1.035 & 51.400 & 1.63 ± 1.024 & 53.230 & 3.48 ± 1.790 \\
\bottomrule
\end{tabular}
\vspace{0.1cm}
\end{minipage}\hfill
\begin{minipage}[t]{\textwidth}
\centering
\subcaption{TD3}
\begin{tabular}{l c cc cc cc c}
\toprule
 & \multirow{2}{*}{\texttt{climlab}} & \multicolumn{2}{c}{\texttt{fed05}} & \multicolumn{2}{c}{\texttt{fed10}} & \multicolumn{2}{c}{\texttt{nofed}} & \multirow{2}{*}{\texttt{ebm-v1}} \\
 &  & Mean ± Std & Gain \% & Mean ± Std & Gain \% & Mean ± Std & Gain \% &  \\
\midrule
90°S–60°S & 11.453 & 7.53 ± 1.444 & 22.700  & 7.42 ± 1.159 & 23.880  & 7.54 ± 1.390 & 22.630  & 9.74 ± 3.330 \\
60°S–30°S & 7.768  & 5.99 ± 2.663 & 15.800  & 5.51 ± 2.616 & 22.590  & 5.00 ± 2.586 & 29.790  & 7.12 ± 3.800 \\
30°S–0°   & 2.730  & 3.84 ± 2.378 & -1.030  & 3.65 ± 2.640 & 3.990   & 3.32 ± 1.380 & 12.690  & 3.80 ± 1.713 \\
0°–30°N   & 3.746  & 7.52 ± 2.875 & -165.300& 7.54 ± 3.263 & -166.260& 5.94 ± 2.701 & -109.610& 2.83 ± 1.247 \\
30°N–60°N & 6.398  & 6.55 ± 1.837 & -29.630 & 6.85 ± 2.390 & -35.450 & 7.02 ± 2.581 & -38.880 & 5.06 ± 2.602 \\
60°N–90°N & 5.566  & 3.94 ± 2.302 & 27.250  & 4.00 ± 1.708 & 26.240  & 4.49 ± 1.536 & 17.170  & 5.42 ± 2.721 \\
\bottomrule
\end{tabular}
\vspace{0.1cm}
\end{minipage}\hfill
\begin{minipage}[t]{\textwidth}
\centering
\subcaption{TQC}
\begin{tabular}{l c cc cc cc c}
\toprule
 & \multirow{2}{*}{\texttt{climlab}} & \multicolumn{2}{c}{\texttt{fed05}} & \multicolumn{2}{c}{\texttt{\texttt{fed10}}} & \multicolumn{2}{c}{\texttt{nofed}} & \multirow{2}{*}{\texttt{ebm-v1}} \\
 &  & Mean ± Std & Gain \% & Mean ± Std & Gain \% & Mean ± Std & Gain \% &  \\
\midrule
90°S–60°S & 11.453 & 8.27 ± 0.378 & 8.700   & 10.78 ± 4.522 & -18.950  & 8.26 ± 0.303 & 8.840   & 9.06 ± 1.549 \\
60°S–30°S & 7.768  & 7.51 ± 0.869 & 5.780   & 10.76 ± 4.459 & -35.000  & 7.62 ± 0.729 & 4.450   & 7.97 ± 1.459 \\
30°S–0°   & 2.730  & 2.60 ± 0.718 & -15.960 & 7.27 ± 5.706  & -224.680 & 3.02 ± 0.490 & -34.920 & 2.24 ± 0.868 \\
0°–30°N   & 3.746  & 2.84 ± 0.774 & -54.750 & 9.67 ± 12.637 & -426.670 & 3.90 ± 0.434 & -112.690& 1.84 ± 0.561 \\
30°N–60°N & 6.398  & 3.59 ± 1.179 & -48.070 & 11.05 ± 17.379& -355.740 & 4.13 ± 0.461 & -70.350 & 2.42 ± 0.706 \\
60°N–90°N & 5.566  & 3.35 ± 1.003 & -43.330 & 11.17 ± 18.477& -378.590 & 3.36 ± 0.394 & -43.800 & 2.33 ± 0.767 \\
\bottomrule
\end{tabular}
\end{minipage}
\end{table}

\begin{table}[!h]
\centering
\small
\caption{Zonal‑band errors for \texttt{ebm-v3-optim-L-20k-a6}. Each subtable reports mean ± std and relative gain \% versus \texttt{ebm-v1} for three regimes \texttt{fed05}, \texttt{fed10} and \texttt{nofed}, along with a comparison against the static baseline \texttt{climlab}}
\setlength{\tabcolsep}{4pt}
\captionsetup[sub]{font=small}
\begin{minipage}[t]{\textwidth}
\centering
\subcaption{DDPG}
\begin{tabular}{l c cc cc cc c}
\toprule
 & \multirow{2}{*}{\texttt{climlab}} & \multicolumn{2}{c}{\texttt{fed05}} & \multicolumn{2}{c}{\texttt{fed10}} & \multicolumn{2}{c}{\texttt{nofed}} & \multirow{2}{*}{\texttt{ebm-v1}} \\
 &  & Mean ± Std & Gain \% & Mean ± Std & Gain \% & Mean ± Std & Gain \% &  \\
\midrule
90°S–60°S & 11.453 & 7.01 ± 2.782 & 14.340  & 7.22 ± 1.399 & 11.800  & 7.23 ± 1.652 & 11.690  & 8.19 ± 3.433 \\
60°S–30°S & 7.768  & 4.34 ± 3.372 & 31.000  & 3.69 ± 1.102 & 41.450  & 8.08 ± 4.816 & -28.370 & 6.30 ± 4.019 \\
30°S–0°   & 2.730  & 1.25 ± 0.695 & 81.440  & 1.22 ± 0.526 & 81.890  & 19.92 ± 4.845& -194.680& 6.76 ± 3.013 \\
0°–30°N   & 3.746  & 1.48 ± 0.617 & 67.830  & 1.71 ± 1.218 & 62.710  & 17.39 ± 5.346& -278.850& 4.59 ± 2.068 \\
30°N–60°N & 6.398  & 1.51 ± 0.710 & 43.220  & 1.57 ± 0.796 & 40.880  & 5.92 ± 6.272& -123.010& 2.65 ± 0.997 \\
60°N–90°N & 5.566  & 1.17 ± 0.442 & 66.490  & 1.39 ± 0.680 & 60.240  & 1.76 ± 1.176& 49.480 & 3.48 ± 1.790 \\
\bottomrule
\end{tabular}
\vspace{0.1cm}
\end{minipage}\hfill
\begin{minipage}[t]{\textwidth}
\centering
\subcaption{TD3}
\begin{tabular}{l c cc cc cc c}
\toprule
 & \multirow{2}{*}{\texttt{climlab}} & \multicolumn{2}{c}{\texttt{fed05}} & \multicolumn{2}{c}{\texttt{fed10}} & \multicolumn{2}{c}{\texttt{nofed}} & \multirow{2}{*}{\texttt{ebm-v1}} \\
 &  & Mean ± Std & Gain \% & Mean ± Std & Gain \% & Mean ± Std & Gain \% &  \\
\midrule
90°S–60°S & 11.453 & 14.05 ± 2.941 & -44.230 & 15.86 ± 1.632 & -62.780 & 12.47 ± 8.601 & -28.000 & 9.74 ± 3.330 \\
60°S–30°S & 7.768  & 10.90 ± 0.644 & -53.120 & 10.59 ± 0.481 & -48.860 & 9.36 ± 5.302  & -31.570 & 7.12 ± 3.800 \\
30°S–0°   & 2.730  & 21.30 ± 0.639 & -460.730& 21.14 ± 0.473 & -456.380& 17.06 ± 4.570 & -349.090& 3.80 ± 1.713 \\
0°–30°N   & 3.746  & 21.41 ± 0.467 & -655.550& 21.23 ± 0.611 & -649.370& 14.34 ± 2.496 & -405.970& 2.83 ± 1.247 \\
30°N–60°N & 6.398  & 9.47 ± 0.516  & -87.290 & 9.36 ± 0.581  & -85.160 & 5.78 ± 0.977  & -14.340 & 5.06 ± 2.602 \\
60°N–90°N & 5.566  & 4.57 ± 0.484  & 15.600  & 4.48 ± 0.536  & 17.280  & 9.27 ± 6.494  & -71.180 & 5.42 ± 2.721 \\
\bottomrule
\end{tabular}
\vspace{0.1cm}
\end{minipage}\hfill
\begin{minipage}[t]{\textwidth}
\centering
\subcaption{TQC}
\begin{tabular}{l c cc cc cc c}
\toprule
 & \multirow{2}{*}{\texttt{climlab}} & \multicolumn{2}{c}{\texttt{fed05}} & \multicolumn{2}{c}{\texttt{\texttt{fed10}}} & \multicolumn{2}{c}{\texttt{nofed}} & \multirow{2}{*}{\texttt{ebm-v1}} \\
 &  & Mean ± Std & Gain \% & Mean ± Std & Gain \% & Mean ± Std & Gain \% &  \\
\midrule
90°S–60°S & 11.453 & 21.12 ± 1.703 & -133.090 & 23.23 ± 8.282 & -156.370 & 21.01 ± 1.702 & -131.930 & 9.06 ± 1.549 \\
60°S–30°S & 7.768  & 18.84 ± 0.971 & -136.340 & 22.08 ± 8.752 & -177.000 & 18.43 ± 1.047 & -131.180 & 7.97 ± 1.459 \\
30°S–0°   & 2.730  & 9.12 ± 0.779  & -307.220 & 10.84 ± 4.123 & -383.720 & 8.54 ± 0.511  & -281.410 & 2.24 ± 0.868 \\
0°–30°N   & 3.746  & 9.41 ± 0.789  & -412.530 & 10.82 ± 2.018 & -489.610 & 8.62 ± 0.503  & -369.770 & 1.84 ± 0.561 \\
30°N–60°N & 6.398  & 11.87 ± 2.754 & -389.580 & 15.20 ± 5.150 & -527.000 & 9.85 ± 0.519  & -306.400 & 2.42 ± 0.706 \\
60°N–90°N & 5.566  & 14.21 ± 7.349 & -509.050 & 20.12 ± 8.341 & -762.090 & 9.07 ± 1.682  & -288.430 & 2.33 ± 0.767 \\
\bottomrule
\end{tabular}
\end{minipage}
\end{table}

\end{landscape}

\clearpage

\subsection{TD3 and TQC Performance Across FedRL EBM Configurations}
\label{app:td3-tqc-additional-results}

\begin{figure}[htbp]
    \centering
    \begin{subfigure}{0.49\textwidth}
        \includegraphics[width=\linewidth]{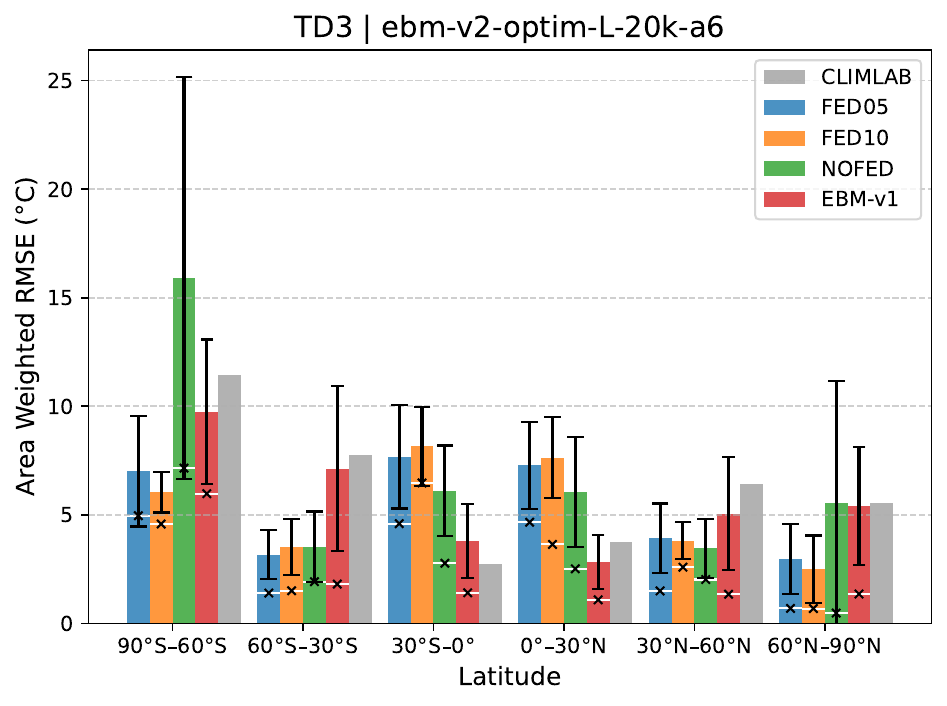}
        \caption{TD3 (\texttt{ebm-v2} \texttt{a6})}
        \label{fig:td3-ebm-v2-a6}
    \end{subfigure}
    \hfill
    \begin{subfigure}{0.49\textwidth}
        \includegraphics[width=\linewidth]{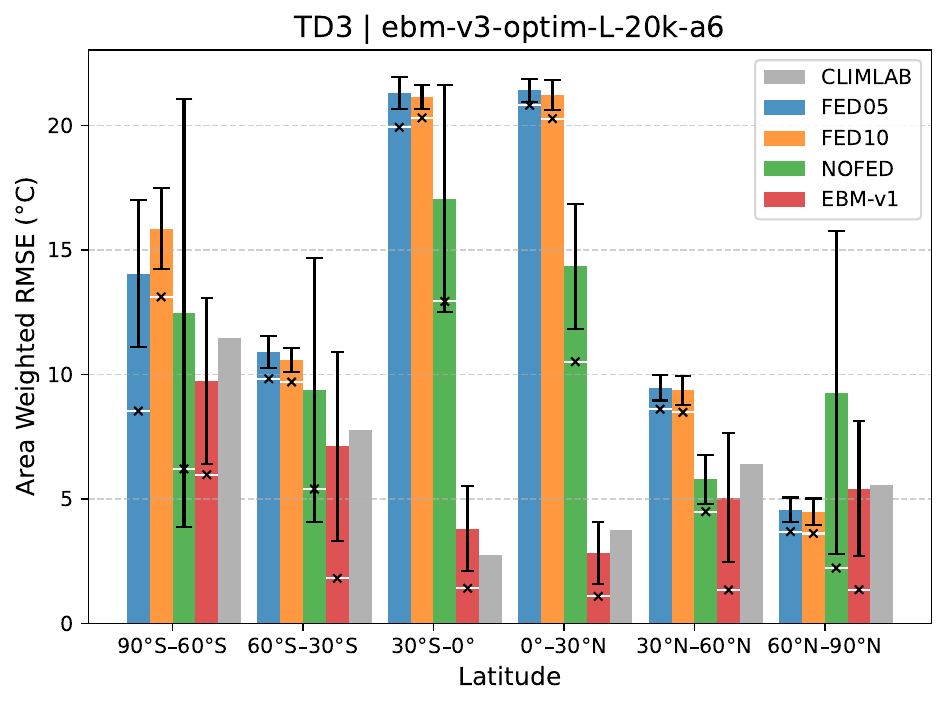}
        \caption{TD3 (\texttt{ebm-v3} \texttt{a6})}
        \label{fig:td3-ebm-v3-a6}
    \end{subfigure}
    \begin{subfigure}{0.49\textwidth}
        \includegraphics[width=\linewidth]{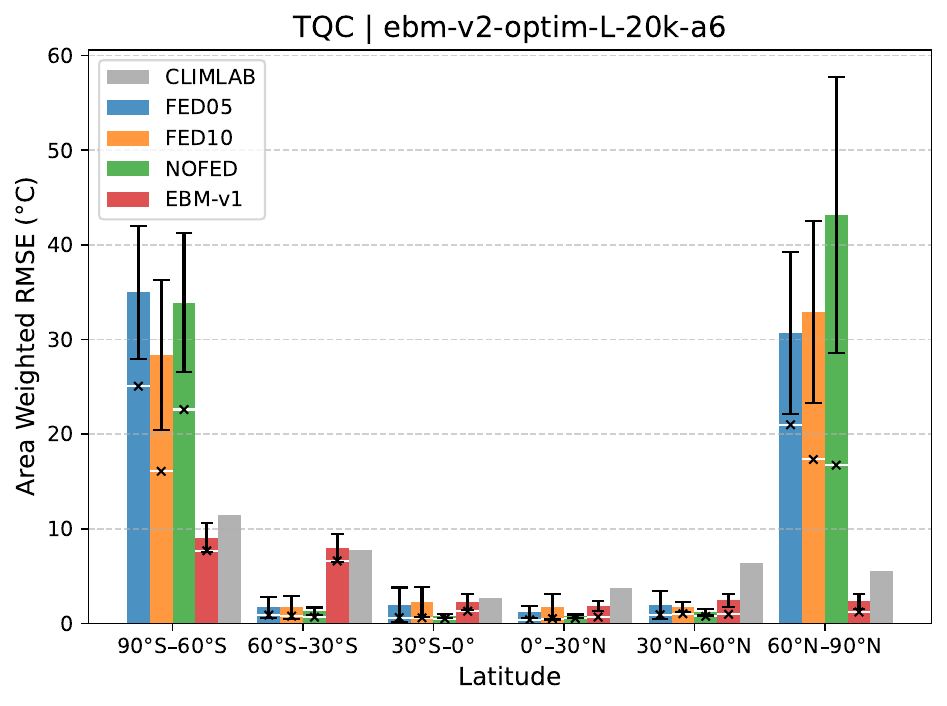}
        \caption{TQC (\texttt{ebm-v2} \texttt{a6})}
        \label{fig:tqc-ebm-v2-a6}
    \end{subfigure}
    \hfill
    \begin{subfigure}{0.49\textwidth}
        \includegraphics[width=\linewidth]{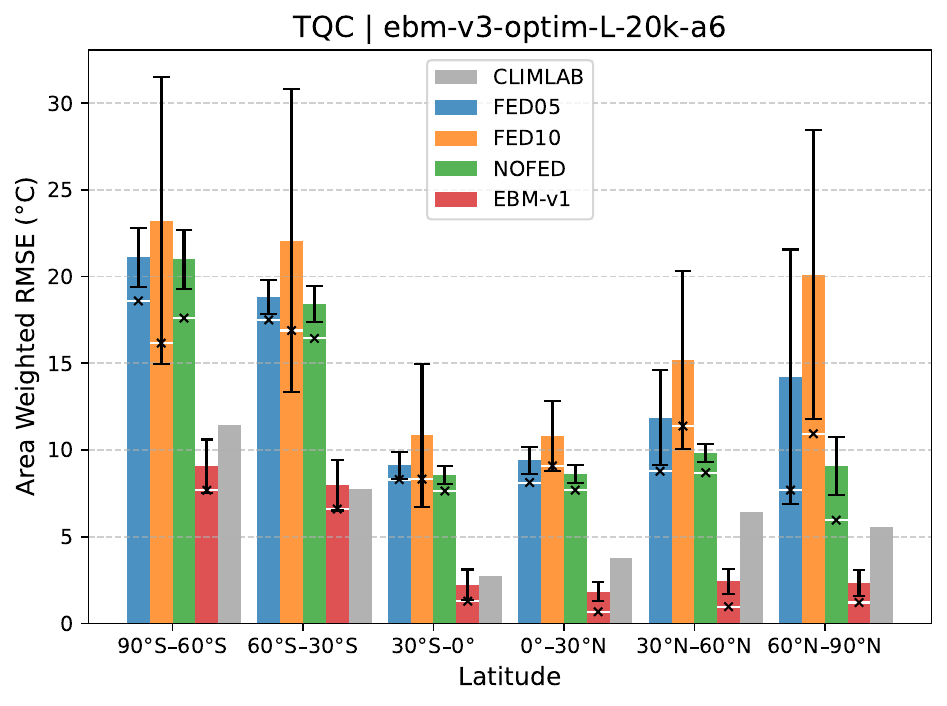}
        \caption{TQC (\texttt{ebm-v3} \texttt{a6})}
        \label{fig:tqc-ebm-v2-a6}
    \end{subfigure}    
    \caption{Comparison of zonal skill (areaWRMSE) for TD3 and TQC across \texttt{ebm-v2} and \texttt{ebm-v3} in 6-agent federated setups averaged with 95\% spreads over 10 seeds. White horizontal bars with a cross indicate the best-performing seed for each algorithm. Both setups adopt the same policy architecture and hyperparameters as \texttt{ebm-v1}. Skill metrics are presented in Appendix~\ref{app:ebm-skill-metrics}.}
    \label{fig:td3-tqc-ebm-v2-v3}
\end{figure}

For TD3 (in Figures~\ref{fig:td3-tqc-ebm-v2-v3} (a) and (b)), performance in \texttt{ebm-v2} shows competitive skill in tropical and mid-latitude zones under \texttt{fed05}, often matching or exceeding \texttt{ebm-v1}. However, variance increases substantially in the polar bands, particularly in the Southern Hemisphere, where sharp gradients appear harder to capture. In \texttt{ebm-v3}, TD3 displays more pronounced instability. While \texttt{fed05} remains the most stable regime, episodic collapses in high-latitude bands lead to elevated RMSE compared to \texttt{ebm-v2}. This suggests that the reduced and region-specific input state in \texttt{ebm-v3}, may cause mismatch between hyperparameters tuned for global state inputs (in \texttt{ebm-v1}) and the regional profile inputs used in \texttt{ebm-v3}, leading to instability in critic ensemble updates.

TQC (in Figures~\ref{fig:td3-tqc-ebm-v2-v3} (c) and (d)) performs strongly in \texttt{ebm-v2} tropical bands under \texttt{fed05}, with clear gains over \texttt{ebm-v1}. However, the method shows instability in high-latitude zones and wider error bars under \texttt{fed10}, indicating a reliance on more frequent synchronisation for stability. In \texttt{ebm-v3}, TQC’s performance degrades notably in the mid-latitudes, with polar areaWRMSE exceeding that of \texttt{ebm-v1} in several cases. The large critic ensemble, which benefits global contexts, may be less effective when regional input profiles and rewards dominate, leading to overfitting or noisy updates. These patterns reaffirm that while both TD3 and TQC can yield strong results under favourable settings, DDPG’s simpler architecture appears more robust to the structural changes between \texttt{ebm-v2} and \texttt{ebm-v3}.

\clearpage

\section{Algorithm Pseudocode}
\label{app:algorithm-pseudocode}

\subsection{Deep Deterministic Policy Gradient (DDPG)}

\begin{algorithm}[!h]
\caption{Deep Deterministic Policy Gradient (DDPG)}
\label{alg:app-pseudocode-DDPG}
\begin{algorithmic}[1]

\State \textbf{Input:} Gym environment, Total timesteps $T$, Replay buffer size $N$, Discount factor $\gamma$, Target smoothing coefficient $\tau$, Batch size $B$, Learning rate $\eta$, Exploration noise $\sigma$

\State \textbf{Initialise:} Policy network parameters $\theta$, Q-function network parameters $\phi$, target network parameters $\theta_{\text{targ}}$, $\phi_{\text{targ}}$, empty replay buffer $\mathcal{D}$

\State \textbf{Pre-Setup:} Configure seed and environment variables, prepare environment and logging \\

\For{$t = 1$ \textbf{to} $T$}
    \State Observe state $s$ and select action $a = \pi_{\theta}(s)$ \State Add exploration noise $a \gets a + \epsilon$, where $\epsilon \sim \mathcal{N}(0, \sigma)$ if required
    \State Execute action $a$ and observe next state $s'$, reward $r$, and termination signal $d$
    \State Store transition $(s, a, r, s', d)$ in $\mathcal{D}$
    \If{$t \geq \text{learning\_starts}$}
        \State Sample a minibatch of $B$ transitions $(s, a, r, s', d)$ from $\mathcal{D}$
        \State Compute target for Q-function update:
        \State \[
        y(r, s', d) = r + \gamma (1 - d) Q_{\phi_{\text{targ}}}(s', \pi_{\theta_{\text{targ}}}(s'))
        \]
        \State Update Q-function by minimising the loss:
        \State \[
        \phi \gets \phi - \eta \nabla_{\phi} \frac{1}{|B|} \sum_{(s,a,r,s',d) \in B} \left( Q_{\phi}(s, a) - y(r, s', d) \right)^2
        \]
        \State Update policy by one step of gradient ascent:
        \State \[
        \theta \gets \theta + \eta \nabla_{\theta} \frac{1}{|B|} \sum_{s \in B} Q_{\phi}(s, \pi_{\theta}(s))
        \]
        \State Soft-update target networks:
        \State \[
        \theta_{\text{targ}} \gets \tau \theta + (1 - \tau) \theta_{\text{targ}},  \quad
        \phi_{\text{targ}} \gets \tau \phi + (1 - \tau) \phi_{\text{targ}}
        \]
    \EndIf
\EndFor

\end{algorithmic}
\end{algorithm}

\clearpage

\subsection{Twin Delayed DDPG (TD3)}

\begin{algorithm}[!h]
\caption{Twin Delayed DDPG (TD3)}
\label{alg:app-pseudocode-TD3}
\begin{algorithmic}[1]
\State \textbf{Input:} Gym environment, Total timesteps $T$, Learning rate $\eta$, Replay buffer size $N$, Discount factor $\gamma$, Target smoothing coefficient $\tau$, Batch size $B$, Policy noise $\sigma_{\pi}$, Noise clip $\sigma_{\text{clip}}$, Exploration noise $\sigma_{\text{exploration}}$, Policy update frequency $f_{\pi}$

\State \textbf{Initialise:} Actor network $\theta$, Critic networks $\phi_1$, $\phi_2$, Target networks $\theta_{targ}$, $\phi_{targ, 1}$, $\phi_{targ, 2}$, Empty replay buffer $\mathcal{D}$

\State \textbf{Pre-Setup:} Configure seed and environment variables, prepare environment and logging \\

\For{$t = 1$ \textbf{to} $T$}
    \State Observe state $s$ and select action $a = \pi_{\theta}(s)$ \State Add exploration noise $a \gets a + \epsilon$, where $\epsilon \sim \mathcal{N}(0, \sigma_{\text{exploration}})$ if required
    \State Execute action $a$ and observe next state $s'$, reward $r$, and done signal $d$
    \State Store transition $(s, a, r, s', d)$ in $\mathcal{D}$
    \If{$t \geq \text{learning\_starts}$}
        \State Sample a minibatch of $B$ transitions $(s, a, r, s', d)$ from $\mathcal{D}$
        \State Compute target actions:
        \State \[
        a' \leftarrow \pi_{\theta_{targ}}(s') + \text{clip}(\mathcal{N}(0, \sigma_{\pi}), -\sigma_{\text{clip}}, \sigma_{\text{clip}})
        \]
        \State Compute target Q-values:
        \State \[
        y(r, s', d) \leftarrow r + \gamma (1 - d) \min_{i=1,2} Q_{\phi_{targ, i}}(s', a')
        \]
        \State Update critic networks by minimising the loss:
        \State \[
        \quad \phi_i \leftarrow \phi_i - \eta \nabla_{\phi_i} \frac{1}{|B|} \sum_{(s,a,r,s',d) \in B} \left( Q_{\phi_i}(s, a) - y(r, s', d) \right)^2, \text{for } i = 1, 2 
        \]
        \If{$t \mod f_{\pi} = 0$}
            \State Update actor network by policy gradient:
            \State \[
            \theta \leftarrow \theta + \eta \nabla_{\theta} \frac{1}{|B|} \sum_{s \in B} Q_{\phi_1}(s, \pi_\theta(s))
            \]
            \State Soft update target networks:
            \State \[
            \theta_{targ} \leftarrow \tau \theta + (1 - \tau) \theta_{targ}, \quad \phi_{targ, i} \leftarrow \tau \phi_i + (1 - \tau) \phi_{targ, i} \text{ for } i=1,2
            \]
        \EndIf
    \EndIf
\EndFor
\end{algorithmic}
\end{algorithm}

\clearpage

\subsection{Truncated Quantile Critics (TQC)}

\enlargethispage{2\baselineskip}

\begin{algorithm}[!h]
\caption{Truncated Quantile Critics (TQC)}
\label{alg:app-pseudocode-TQC}
\begin{algorithmic}[1]
\State \textbf{Input:} Gym environment, Total timesteps $T$, Replay buffer size $N$, Discount factor $\gamma$, Smoothing coefficient $\tau$, Batch size $B$, Learning rate $\eta$, Number of quantiles $N_q$, Number of critics $N_c$, Drop quantiles $N_{\text{drop}}$, Entropy coefficient $\alpha$, Target entropy coefficient $\alpha_{\text{targ}}$

\State \textbf{Initialise:} Actor network $\theta$, Critic network parameters $\phi_1, \dots, \phi_{N_c}$, Target critic network parameters $\phi_{\text{targ},1}, \dots, \phi_{\text{targ},N_c}$, Replay buffer $\mathcal{D}$

\State \textbf{Pre-Setup:} Configure seed and environment variables, prepare environment and logging \\

\For{$t = 1$ \textbf{to} $T$}
    \State Select action $a \sim \pi_\theta(s)$ based on current policy and exploration strategy
    \State Execute action $a$ and observe next state $s'$, reward $r$, and done signal $d$
    \State Store transition tuple $(s, a, r, s', d)$ in $\mathcal{D}$
    \If{$t \geq \text{learning\_starts}$} 
        \For{$i = 1$ \textbf{to} $N_c$}
            \State Sample a minibatch of $B$ transitions $(s, a, r, s', d)$ from $\mathcal{D}$
            \State Compute target quantile values for critic $\phi_{target, i}$:
            \State \[
        y(r, s', d) = r + \gamma (1 - d) \left( Q_{\phi_{\text{targ},i}}(s', \tilde{a}', N_{\text{drop}}) - \alpha \log \pi_\theta(\tilde{a}' | s') \right) \]
        \State where $\tilde{a}' \sim \pi_\theta(s')$        
            \State Update critic $\phi_i$ by minimising the quantile Huber loss:
            \[
            \text{L}^{\phi_i} = \frac{1}{N_q} \sum_{k=1}^{N_q} \text{HuberLoss}(Q_{\phi_i}(s_j, a_j, \tau_k) - y_j) \]
            \State where $\tau_k$ are the quantile fractions
        \EndFor
        \State Update policy by one step of gradient ascent:
        \State \[
        \theta \gets \theta + \eta \nabla_\theta \frac{1}{|B|} \sum_{s \in B} \left( - \alpha \log \pi_\theta(a|s) + \frac{1}{N_c} \sum_{i=1}^{N_c} Q_{\phi_i}(s, \pi_\theta(s)) \right)
        \]
        \State Soft-update target networks:
        \State \[
        \phi_{\text{targ},i} \gets \tau \phi_i + (1 - \tau) \phi_{\text{targ},i} \text{ for } i = 1, 2, ..., N_c
        \]
        \State Optionally adjust $\alpha$ based on entropy targets:
        \State \[
        \alpha \gets \alpha + \eta \nabla_{\alpha} \frac{\alpha}{|B|}\sum_{s \in B} \left( \log \pi_\theta(a|s) + \alpha_{\text{targ}} \right)
        \]
    \EndIf
\EndFor

\end{algorithmic}
\end{algorithm}

\end{document}